\documentclass[10pt,a4paper,oneside]{article}
\usepackage[utf8]{inputenc}

\newcommand{\mc}{\mathcal}
\newcommand{\mb}{\mathbb}
\usepackage{hyperref}
\usepackage{url}
\usepackage{lipsum}
\usepackage{amsmath}
\usepackage{amssymb}
\usepackage[round]{natbib}
\usepackage{bm}
\newcommand{\uset}[2]{\underset{#1}{#2}}
\DeclareMathOperator*{\argmax}{arg\,max}

\begin{document}
\title{A Succinct Summary of Reinforcement Learning}
\author{Sanjeevan Ahilan\footnote{sanjeevanahilan@gmail.com. Much of this work was done at the Gatsby Unit, UCL.}} 
\date{}

\maketitle

\begin{abstract}This document is a concise summary of many key results in single-agent reinforcement learning (RL). The intended audience are those who already have some familiarity with RL and are looking to review, reference and/or remind themselves of important ideas in the field.\end{abstract}

 \tableofcontents

 \newpage

\section{Acknowledgements}
I would like to thank Peter Dayan, David Silver, Chris Watkins and ChatGPT for helpful feedback. Much of this work was drawn from David Silver's UCL course\footnote{\url{https://www.davidsilver.uk/teaching/}} and Sutton and Barto's textbook \citep{sutton2018reinforcement} and formed the introductory chapter of my PhD thesis \citep{ahilan2021structures}. 

% \footnote{our review draws from David Silver's UCL course available at\url{http://www0.cs.ucl.ac.uk/staff/d.silver/web/Teaching.html} and Sutton and Barto's textbook \citep{sutton2018reinforcement}}. We then describe model-free RL, introducing algorithms 
% in deep RL relevant to later chapters. Finally, we introduce latent variable models and the problem of
% partial observability.
\section{Fundamentals} 

\subsection{The RL paradigm}

%Reinforcement learning is a field in machine learning in which an
%agent takes actions in an environment which influences the subsequent data it receives. The agent optimises an
%objective which is a notion of cumulative reward, where reward is a scalar signal. The problem of reinforcement
%learning is that of decision-making; learning to select actions which maximise the total reward.

% The beginning of an animal's life is often characterised by weakness, a lack of skill and
% reliance on parents to provide sustenance. Given time and repeated interactions with their environment, however,
% animals will generally acquire the necessary skills for survival, whether this be hunting prey or navigating complex
% terrain. 

% Rather than directly theorising about how people or animals learn, it
% explores idealised learning situations and evaluates the performance of various learning methods.

The field of reinforcement learning (RL) \citep{sutton2018reinforcement} concerns itself with the computational principles underlying goal-directed learning through interaction. 
Although primarily seen as a field of machine learning, it has a rich history spanning multiple fields. In psychology it can be used to model classical (Pavlovian) and operant (instrumental) conditioning. In neuroscience it has been used to model the dopamine system of the brain \citep{schultz1997neural}. In economics, it relates to fields such as bounded rationality, and in engineering it has extensive overlap with the field of optimal control \citep{bellman1957markovian}. In mathematics,
investigation has continued under the guise of operations research. The plethora of perspectives ensures that RL continues to be an exciting and extraordinarily
interdisciplinary field.

\subsection{Agent and environment}
RL problems typically draw a separation between the agent and the
environment. The agent receives observation $o_t$ and scalar reward $r_t$ from the environment and emits action $a_t$, 
where $t$ 
indicates the time step. The environment receives action $a_t$ from the agent and then emits a reward $r_{t+1}$ 
and an observation $o_{t+1}$. The cycle then begins again with the agent emitting its next action.

How the environment responds to the agent's action is determined by the environment state $s_{t}$, which is updated at 
every time step. The conditional distribution for the next environment state depends only on the present state and 
action and therefore satisfies the Markov property:

%The agent emits action $a_t$, receives observation $o_t$,
%and scalar reward $r_t$,  Correspondingly,
%the environment, receives action $a_t$, emits observation $o_t$ and
%scalar reward $r_t$. The environment state  
\begin{equation}
P(s_{t+1} | s_{t}, a_{t}) = P(s_{t+1} | s_{1},\ldots, s_{t}, a_{1},\ldots, a_{t})
\end{equation}

The environment state is in general private from the agent, which only receives observations and rewards. The 
conditional distribution for the next observation given the current observation is not in general Markov, and so it may 
be beneficial for an agent to construct its own notion of state $s^\alpha_t$, which it uses to determine its next action. 
This can be defined as $s^\alpha_{t} = f(h_{t})$, where $h_t$ is the history of the agent's sequence of observations, 
actions and rewards: 
%\\\tope{so  you make the environment Markov -- perhaps say SA: 
%difficult to say until after the definition, I have added it at the beginning of the next subsection}
\begin{equation}
h_{t} = a_{1}, o_{1}, r_{1},\ldots,a_{t}, o_{t}, r_{t}
\end{equation}

%In general the conditional distribution for the next observation given the prevAn agent may wish to construct its own 
%notion of state $s^a_t$ which is Markov in 
%Transitions according to the own agent's state may or may not satisfy the Markov property. 

%The conditional distribution for the next state satisfies the Markov property if it depends only on the current 
%state:
%-- the state is a sufficient statistic of the
%future.
%\begin{equation}
%P(s_{t+1} | s_{t}) = P(s_{t+1} | s_{1},\ldots, s_{t})
%\end{equation}
%
%In other words, the future is independent of the past given the present - the state is a sufficient statistic of the
%future. 
%\tope{meaning that $o$ includes $s^e$ and that you can replace $h_t$
%	just by $s^e_t$? SA: yes} 

\subsection{Observability}
A special case exists when the observation received by the agent $o_t$ is identical to the environment state $s_t$ 
(such that there is no need to distinguish between the two). This is the assumption underlying the formalism of Markov 
decision processes covered in the next section. An environment is partially observable if the agent cannot observe the 
full environment state, meaning that the conditional distribution for its next observation given its current observation
does not satisfy the Markov property. This assumption underlies the
formalism of a partially observable Markov decision process which we describe in Section \ref{background:pomdp}.  

\subsection{Markov processes and Markov reward processes}
A Markov process (or Markov chain) is a sequence of random states with the Markov property. It is defined in terms of
the tuple $\langle \mc{S}, \mc{P} \rangle$ where $\mc{S}$ is a finite set of states and $\mc{P} : \mc{S} \times
\mc{S} \rightarrow [0,1]$ is the state transition probability
kernel. 

A Markov Reward Process (MRP) $\langle \mc{S}, \mc{P}, r, \gamma
\rangle$ extends the Markov process by including a reward function $r:
\mc{S}  \times \mc{S} \rightarrow \mb{R}$ for each state transition and a discount factor $\gamma$. The immediate 
expected reward in a given state is defined as: $r(s) = \sum_{s'} \mc{P}(s, s') r(s, s')$.

The discount factor $\gamma \in [0,1]$ is used
to determine the present value of future rewards. Conventionally, a reward received $k$
steps into the future is of worth $\gamma^k$ times what it would be
worth if received immediately. As we will shortly see, the cumulative sum of 
discounted rewards is a quantity RL agents often seek to maximise, and so
$\gamma < 1$ ensures that this sum is bounded (assuming $r$ is bounded). 
%Lower 
%values of $\gamma$
%encourage agents to be more short sighted whereas higher values
%encourage far sighted behaviour.

\subsection{Markov decision processes}
Single-agent RL can be formalised in terms of Markov decision processes
(MDPs). The idea of an MDP is to capture the key components available to
the learning agent; the agent's sensation of the state of its
environment, the actions it takes which can affect the state, and the
rewards associated with states and actions. An MDP extends the formalism
of an MRP to include a finite set of actions on which both $\mc{P}$ and
$r$ depend.
%An MDP assumes the
%Markov Property such that the outcome of an action in a state is independent of past states and actions.
Discrete-time, infinite-horizon MDPs are described in terms of the 5-tuple $\langle \mc{S}$, $\mc{A}$, $\mc{P}$, $r$, 
$\gamma
\rangle$ where $\mc{S}$ is the set of states, $\mc{A}$ is the set of actions, $\mc{P} : \mc{S} \times \mc{A} \times \mc{
	S} \rightarrow [0,1]$ is the state transition probability kernel, $r: \mc{S} \times \mc{A} \times \mc{S} 
	\rightarrow \mb{R}$ is the immediate reward function and $\gamma \in [0,1)$ is the discount factor. The expected 
	immediate reward for a given state and action is defined as $r(s, a) = \sum_{s'} \mc{P}(s, a, s') r(s, a, s')$, 
	which we use for convenience subsequently. 

\subsection{Policies, values and models} Common components of a reinforcement learning agent are a policy, value
function and a model. The policy $\pi : \mc{S} \times \mc{A} \rightarrow [0,1]$ is the agent's behaviour function which
denotes the probability of taking action $a$ in state $s$. Agents may also act according to a deterministic policy $
\mu: \mc{S} \rightarrow \mc{A}$. We will assume that policies are stochastic unless otherwise noted.

Given an MDP and a policy $\pi$, the observed state sequence is a Markov process $\langle \mc{S},
\mc{P}^{\pi} \rangle$. 

\begin{equation}
\mc{P}^{\pi}(s,s') = \sum_{a \in \mc{A}} \pi(s, a)\mc{P}(s,a,s')
\end{equation}

Similarly, the state and reward sequence is a MRP $\langle \mc{S}, \mc{P}^{\pi}, r_{\pi},
\gamma \rangle$ in which:

\begin{equation}
%r_{\pi}(s, s') = \sum_{a \in \mc{A}} \pi(s,a)\mc{P}(s,a,s')r(s,a,s')
r_{\pi}(s) = \sum_{a \in \mc{A}} \pi(s,a)r(s,a)
\end{equation}

Starting from any particular state $s$ at time step $t=0$, the value function $v_{\pi}(s)$ is a prediction of the
expected discounted future reward given that the agent starts in state $s$ and follows policy $\pi$:

\begin{equation}
\label{eq:value}
v_{\pi}(s) = \mb{E}_{\pi} \Bigg[\sum_{t=0}^{\infty} \gamma^t r_{t+1}|s_0 = s \Bigg]
\end{equation}
where $r_{t+1}=r(s_t,a_t,s_{t+1})$

which is the solution of an associated Bellman expectation equation:

\begin{equation}
\label{eq:bellman_expectation}
v_{\pi}(s) = \sum_{a \in A} \pi(s,a) \Bigg[ r(s,a)+ \gamma \sum_{s' \in S} \mc{P}(s,a,s')v_\pi (s')\Bigg]
\end{equation}

In matrix form the Bellman expectation equation can be expressed in terms of the induced MRP:

\begin{equation}
\bm{v}_{\pi} = \bm{r}_{\pi}+ \gamma \mc{P}^{\pi}\bm{v_{\pi}} = (\mc{I} - \gamma 
\mc{P}^{\pi})^{-1}
\bm{r}_{\pi}
\end{equation}

where $\bm{v}_{\pi} \in \mb{R}^{|\mc{S}|}$ and $\bm{r}_\pi \in \mb{R}^{|\mc{S}|}$ are the vector of values and expected 
immediate rewards respectively for each state under policy $\pi$. We can also define a Bellman expectation 
backup operator:
\begin{equation}
\label{eq:bellman_expectation_backup}
T^\pi(\bm{v}) = \bm{r}_{\pi} + \gamma \mc{P}^{\pi}\bm{v}
\end{equation}
which has a fixed point of $\bm{v}^\pi$.

An action-value for a policy $\pi$ can also be defined, which is the expected discounted future reward for
executing action $a$ and subsequently following policy $\pi$.

\begin{equation}
\begin{aligned}
q_{\pi}(s,a) & = r(s,a)+ \gamma \sum_{s' \in S} \mc{P}(s,a,s') v_\pi(s')                                     \\
& = r(s,a)+ \gamma \sum_{s' \in S} \mc{P}(s,a,s') \sum_{a' \in \mc{A}} \pi(s',a') q_\pi (s',a')
\end{aligned}
\end{equation}

The process of estimating $v_\pi$ or $q_\pi$ is known as policy
evaluation. Policies can be evaluated without directly knowing or
estimating a model, using instead the directly sampled experience of the
environment, an approach which is known as `model-free'. However a
`model-based' approach is also possible in which a model is used to
predict what the environment will do next. A key component of a model is
an estimate of $ \mc{P}(s,a,s')$, the probability of the next state
given the current state and action. Another is an estimate of $r(s,a)$,
the expected immediate reward.

Policy evaluation enables a value function to be learned for a given policy. However, we often wish to learn the
best possible policy. The value function for this is known as the optimal value function and corresponds to the
maximum value function over all policies:

\begin{equation}
\begin{aligned} v_{*}(s) & = \uset{\pi}{\max} \  v_{\pi}(
s)                                        \\
\end{aligned}
\end{equation}

The definition of the optimal action-value function (which evaluates the immediate action $a$ in state $s$) is 
similarly:

\begin{equation}
\begin{aligned} q_{*}(s,a) & = \uset{\pi}{\max} \  q_{\pi}(s,a)
\end{aligned}
\end{equation}

A partial ordering over policies can be defined according to:
\begin{equation}
\pi \geq \pi' \text{ if } v_{\pi}(s) \geq v_{\pi'}(s), \forall s
\end{equation}

For any MDP there exists an optimal policy $\pi_{*}$ that is better than or equal to all other policies. All optimal
policies achieve the optimal value function and optimal action-value function and there is always a deterministic
optimal policy for any MDP. The latter is achieved by selecting:

\begin{equation}
\label{eq:optimal_action}
a = \uset{a \in \mc{A}}{\argmax} \ q_{*}(s,a)
\end{equation}

If there are many possible actions which satisfy this, any of these may be chosen to constitute an optimal policy (of
which there may be many). The optimal value and state-value functions satisfy Bellman optimality equations:
\begin{equation}
\begin{aligned}
\label{eq:optimal}
v_{*}(s) & = \uset{
	a \in \mc{A}}{\max} \ q_{*}(s,a) \\ v_{*}(s) &= \uset{a \in \mc{A}}{\max} \Big{[}  r(s,a) +  \gamma \sum_{s' \in 
	\mc{S}}
\mc{P}(s,a,s')v_{*}(s')\Big{]}   \\ q_{*}(s,a) &=  r(s,a) + \gamma \sum_{s' \in \mc{S}}\mc{P}(s,a,s') 
\uset{a'}{\max} \
q_{*}(s',a')                     \\
\end{aligned}
\end{equation}

The Bellman optimality equation is non-linear with no closed form solution (in general). Solving it therefore
requires  iterative solution methods.

\subsection{Dynamic programming}
% It requires that there be optimal substructure i.e. the optimal solution
%can be constructed from optimal solutions to its subproblems. 
Dynamic programming (DP) \citep{bertsekas1995dynamic} refers to a collection of algorithms that can be used to compute 
optimal policies given a perfect model of the environment as an MDP. In general, DP solves complex problems by
breaking them down into subproblems and then combining the
solutions. It is particularly useful for overlapping subproblems, the solutions to which
reoccur many times when solving the overall problem, making it more
computationally efficient to cache and reuse them.
%refers to a colection of algorithms that can be used to compute 
%optimal policies given a perfect model of the environment as a Markov Decision Process.

When applied to MDPs, the recursive decomposition of DP corresponds to the Bellman equation and the cached solution to 
the value function. DP assumes that the MDP is fully known and therefore
does not address the full RL problem but instead addresses the problem
of planning. By planning, the prediction problem can be addressed by
finding the value function $v_{\pi}$ of a given policy $\pi$. This can
be evaluated by iterative application of the Bellman Expectation Backup
(Equation \ref{eq:bellman_expectation_backup}). 

This leads to convergence to a unique fixed point $v_{\pi}$, which can be shown using the contraction mapping
theorem (also known as the Banach fixed-point theorem) \citep{banach1922operations}. When a Bellman expectation
backup operator $T^{\pi}$ is applied to two value functions $\bm{u}$ and
$\bm{v}$ over states, we find that it is a $\gamma$-contraction: 
%\begin{equation}
%    \max_{s \in \mc{S}} |B\hat{v}(s)-v_{*}(s)| \leq \gamma \max_{s \in \mc{S}} |\hat{v}(s)-v_{*}(s)|
%\end{equation}

\begin{equation}
\begin{aligned}
||T^{\pi}(\bm{u})-T^{\pi}(\bm{v})||_{\infty} & = ||(r_{\pi}+\gamma \mc{P}^{\pi} \bm{u})-
(r_{\pi}+ \gamma \mc{P}^{\pi} \bm{v})||_{\infty}                                          \\
& = ||\gamma \mc{P}^{\pi} (\bm{u} -\bm{v})||_{\infty} \\
& \leq  || \gamma \mc{P}^{\pi}\bm{1}||\bm{u} -\bm{v}||_{\infty}  ||_{\infty} \\
& \leq \gamma||\bm{u} -\bm{v}||_{\infty} \\
\end{aligned}
\end{equation}

where $\bm{1}$ is a vector of ones and the infinity norm of a vector $\bm{a}$ is denoted $||\bm{a}||_{\infty}$ and is 
defined as the maximum value of its components. This contraction ensures that both $\bm{u}$ and $\bm{v}$ converge to
the unique fixed point of $T^{\pi}$ which is $\bm{v}_\pi$.

For control, DP can be used to find the optimal value function $v_{*}$ and in turn the optimal policy $\pi_{*}$. One
possibility is \textit{policy iteration} in which the current policy $\pi$ is first evaluated as described and then
subsequently improved to $\pi'$ such that:

\begin{equation}
\label{eqn:greedyq}
\pi'(s) = \uset{a \in \mc{A}}{\argmax} \ q_{\pi}(s,a)
\end{equation}

This improves the value from any state $s$ over one step:

\begin{equation}
q_{\pi}(s, \pi'(s)) = \uset{a \in \mc{A}}{\max} \ q_{\pi}(s,a) \geq  \sum_{a \in \mc{A}} \pi(s,a) q_\pi (s,a) = 
v_{\pi}(s)
\end{equation}
%\geq q_{\pi}(s,\pi(s))

It can be shown that this improves the value function such that that $v_{\pi'}(s) \geq v_{\pi}(s)$ \citep{silver2015lecture}. This process
is then repeated, with improvements ending when the Bellman optimality
equation (\ref{eq:optimal}) has been satisfied and convergence to
$\pi_*$ achieved. A generalisation of policy iteration is also possible
in which, instead of waiting for policy evaluation to converge, only $n$
steps of evaluation are taken before policy improvement occurs and the
process is repeated. If $n=1$ this is known as \textit{value iteration},
as the policy is no longer explicit (being a direct consequence of the
value function). Like policy iteration, value iteration is also
guaranteed to converge to the optimal value function and policy. This
can be demonstrated using the contraction mapping theorem.

\section{Model-free approaches}
\subsection{Prediction }

As has been outlined, dynamic programming can be used to solve known
MDPs enabling optimal value functions and policies to be found. However,
in many cases the MDP is not directly known - instead an agent taking
actions in the MDP must learn directly from its experiences, as it
transitions from state to state and receives rewards accordingly. One
approach, known as `model-free', seeks to solve MDPs without learning
transitions or rewards. For prediction, a key quantity to estimate in
this setting is the expected discounted future reward. A sampled estimate of this, starting from state $s_t$, is known 
as the return:

\begin{equation} \label{eqn:return}
R_{t} = r_{t+1} + \gamma r_{t+2} + \gamma^2 r_{t+3} + ... = \sum^{\infty}_{k = 0} \gamma^{k}r_{t+k+1}
\end{equation}

which depends on the actions sampled from the policy, and states from transitions.

\textit{Monte-Carlo} (MC) methods seek to estimate this directly using
complete episodes of experience. Introducing a learning rate $\alpha_t$,
the agent's value function can therefore be updated according
to\footnote{assuming a table-based representation rather than use of a
	function approximator}: 
%\topd{what happened to the policy decoration for $v$? SA: my intention was to denote the 
	%true value function for policy $\pi$ as $v_\pi$ whereas this is just an estimate. Does that make sense?}

\begin{equation} v(s_{t}) \leftarrow v(s_{t}) +
\alpha_t \big{[}R_{t}-v(s_t)\big{]}
\end{equation}

The value function updated in this way will converge to a solution with
minimum mean-square error (best fit to the observed returns), assuming a
suitable sequential decrease in the learning rate. 

\textit{Temporal-difference} (TD) learning methods learn from incomplete episodes by bootstrapping. For example, if
learning occurs after a single step, this is known as TD(0), which has
the following update:
\begin{equation} v(s_{t})
\leftarrow v(s_{t}) + \alpha_t \big{[}r_{t+1}+\gamma v(s_{t+1})-v(s_t)\big{]}
\end{equation} where $r_{t+1}+\gamma v(s_{t+1})$ is
known as the \textit{target}. This approximates the full-width Bellman expectation backup (Equation 
\ref{eq:bellman_expectation_backup})
in which every successor state and action is considered, with
experiences instead being sampled. TD(0) will converge to the solution
of the maximum likelihood Markov model which best fits the data (again
assuming a suitable sequential decrease in the learning rate). This
solution may be different from the minimum mean-square error solution of
MC methods, which do not assume the Markov property.

Unlike MC methods, TD methods introduce bias into the estimated return
as the currently estimated value function may be different from the true
value function. However, they generally have reduced variance relative
to MC methods, as in MC the estimated return depends on a potentially
long sequence of random actions, transitions and rewards.

The distinction between MC and TD methods can be blurred by considering multi-step TD methods (rather than only TD(0)),
in which rewards are sampled for a number of steps before the value function is used to compute an estimate of future
rewards. The $n$-step return is defined as:
\begin{equation}
R_{t}^{(n)} = r_{t+1}+\gamma r_{t+2}+...+\gamma^{n-1}r_{t+n}+\gamma^{n} v(s_{t+n})
\end{equation}

As $n \rightarrow \infty$ it tends towards the unbiased MC return. An algorithm may seek to find a good bias-variance 
tradeoff by estimating a weighted combination of n-step returns; one popular method to do this is known as 
TD($\lambda$):

\begin{equation}
R_{t}^{\lambda} = (1-\lambda)\sum_{n=1}^{\infty} \lambda^{n-1} R_{t}^{(n)}
\end{equation}

where $\lambda \in [0,1]$. 

\subsection{Control with action-value functions}

Model free control concerns itself with optimising rather than evaluating the RL objective. Policies may be evaluated
according to various objectives. In the case of continuing environments, the objective can be the average value or
the average reward per time-step. We focus instead on episodic environments, assuming an initial distribution over
starting states $p_0(s): \mc{S} \rightarrow [0,1]$. The objective is thus:

\begin{equation}
J(\pi) = \mb{E}_{\pi} \Bigg[\sum_{t=0}^{\infty} \gamma^t r_{t+1}|p_0(s) \Bigg]
\end{equation}

Note that if the domain of the starting state distribution is only over a single starting state, the objective is
simply the value function (Equation \ref{eq:value}) in that starting state. This objective can equivalently be expressed
as:

\begin{equation}
J(\pi) = \mb{E}_{s \sim \rho^\pi, a \sim \pi}[ r(s,a)]
\label{discounted_state_objective}
\end{equation}

where:

\begin{equation}
%	\rho^\pi(s) := \int_{\mc{S}} \sum_{t=0}^{\infty}
%	\gamma^{t}p_0(s')p(s_t =s| s', \pi) ds'
\rho^\pi(s) := \sum_{s'} \sum_{t=0}^{\infty}
\gamma^{t}p(s_t =s| s', \pi) p_0(s')
\label{discounted_state}
\end{equation}
is the improper discounted state distribution induced by policy $\pi$ starting from an initial state distribution 
$p_0(s')$. In Section \ref{policy_gradients} we describe policy gradient methods which seek to optimise this objective directly.

However, we first consider model-free approaches which rely on an
action-value function $q(s,a)$ to achieve control (a value function
$v(s)$ alone is insufficient for model-free control). The optimal
action-value function $q_{*}(s,a)$ must be learned, with MC and TD
methods both viable. Once it has been learned, an optimal policy may be
achieved by selecting the best action in each state (Equation
\ref{eq:optimal_action}).

However, unlike dynamic programming, full-width backups are not used and
so if actions are selected greedily (meaning those with highest
action-values are always chosen) then certain states and actions may
never be correctly evaluated. Model-free RL methods must therefore allow
for enough exploration during learning before ultimately exploiting this
learning to achieve near-optimal cumulative reward. 

One simple approach, known as $\epsilon$-greedy is to take a random action with probability $\epsilon$  but otherwise
act greedily according to the current estimate of the action-value function. The value of $\epsilon$ can be decreased 
with the number of episodes. This can satisfy a condition known
as greedy in the limit of infinite exploration in which all state-action pairs are explored infinitely many times and
the policy converges to the greedy policy.

One popular algorithm for model-free control is known as Q-learning \citep{watkins1992q},
which seeks to learn the optimal action-value function whilst using a policy which also takes exploratory actions (such 
as epsilon greedy). This learning is termed
\textit{off-policy} as the policy used to sample experience is different from the policy being learned (the
optimal policy). 
The resulting update is:

\begin{equation} q(s_{t},a_{t})
\leftarrow q(s_{t},a_{t}) + \alpha \big{[}r_{t+1}+
\gamma \uset{a' \in \mc{A}}{\max} \ q(s_{t+1},a') -q(s_{t},a_{t})\big{]}
\end{equation}

An alternative to off-policy Q-learning is on-policy SARSA \citep{rummery1994line}. This uses the sampled sampled state $s_{t}$, action
$a_{t}$, reward $r_{t+1}$, next state $s_{t+1}$, and next action $a_{t+1}$ for updates\footnote{and also
	gives SARSA its name}:
%\tope{why did you get rid of the text
%	noting that you have to choose
%	$a_{t+1}$ wisely. SA: hmm, I don't remember getting rid of anything? My version control says its been like that 
%	since January. Should I be saying more?}

\begin{equation}
\begin{aligned}
q(s_{t},a_{t}) & \leftarrow q(s_{t},a_{t})  + \alpha (r_{t+1}+\gamma q(s_{t+1},a_{t+1})-q(s_{t},a_{t})) \\
%\epsilon &\leftarrow \frac{1}{k} \\
%\pi &\leftarrow \epsilon \text{-greedy}(q)
\end{aligned}
\end{equation}
%It is important therefore to
%consider a \textit{greedy} policy which selects $\pi(s) = \argmax_{a \in \mc{A}} Q(s,a)$.

 %  One popular objective is
%to maximise the discounted expected future reward, defined as $\mb{E}_{\pi}[\sum_{t=0}^{\infty} \gamma^t r(s_t, a_t)]$
%where the expectation is over the sequence of states and actions which result from policy $\pi$, starting from an
%initial state distribution $\rho_0: \mc{S} \rightarrow [0,1]$ (where $t$ is the time step). %  This objective can be
%equivalently expressed as $\mb{E}_{s \sim \rho^\pi, a \sim \pi}[\mc{R}(s,a)]$, where $\rho^\pi$ is the discounted state
%distribution induced by policy $\pi$ starting from $\rho_0$.

\subsection{Value function approximation}
So far we have assumed a tabular representation of states and actions such
that each state is separately updated. However, in practice we would like value functions and policies to generalise
to new states and actions, and so it is beneficial to use function approximators such as deep neural networks. A common 
approach is to approximate the value function or action-value function:

\begin{equation}
\begin{aligned}
v_w (s) &= \hat{v}(s; w) \approx v_{\pi}(s) \\
q_w (s,a) &= \hat{q}(s, a; w) \approx q_{\pi}(s,a)
\end{aligned}
\end{equation}
where $w$ are the parameters we wish to learn. If we start by assuming we know the true value function $v_\pi$, we can 
define a mean square error between the approximate value function and the true function:
\begin{equation}
\mc{L}(w) = \mb{E}_{\pi}[(v_{\pi}(s) - v_w(s))^{2}] 
\end{equation}

Given a distribution of states $s \sim p(s)$\footnote{we later discuss a method for sampling states}, we can 
minimise this iteratively using stochastic gradient descent:

\begin{equation}
w \leftarrow w+ \alpha (v_{\pi}(s_t) - v_w (s_t))
\nabla_w v_w (s_t)\\
\end{equation}

In reality we can only use a better estimate of $v_\pi$ provided by the sampled reward(s). For example, if we use the 
TD(0) target the update is:

\begin{equation}
w \leftarrow w+ \alpha (r_{t+1}+\gamma v_w (s_{t+1}) - v_w (s_t))
\nabla_w v_w (s_t)\\
\end{equation}

Updates like this are known as `semi-gradient' as the gradient of the value function used to define the target is
ignored.

If we use a linear function approximator $v_w (s) = x(s)^{T}w$ (where features $x(s)$ and $w$ are vectors), 
then we find:
\begin{equation}
\begin{aligned}
w &\leftarrow w+ \alpha (r_{t+1}+\gamma v_w (s_{t+1}) - v_w (s_t))x(s_t)\\
\end{aligned}
\end{equation}

indicating that the linear weights are updated in proportion to the activity of their corresponding features. 
Non-linear function approximators can also be used, but typically have weaker convergence guarantees than linear 
function approximators. Nevertheless, due to their flexibility such approximators have enabled impressive performance 
in a number of challenging domains, such as Atari games \citep{mnih2015human} and
Go \citep{silver2016mastering}. 

\subsection{Policy gradient methods}
\label{policy_gradients}
Parameterised stochastic policies $\pi_\theta$ may be improved using the
\textit{policy gradient theorem} \citep{sutton2000policy}. This can be derived for any of the common RL objectives. To
demonstrate a derivation of this result we use a starting state objective $J(\theta)=v_{\pi_\theta}(s_0)$ with a single 
starting state $s_0$: 
%\begin{equation}
%\begin{aligned}
%\nabla_\theta J(\theta) &= \nabla_\theta  v_{\pi}(s_0) \\
%\frac{\partial J(\theta)}{ \partial \theta} & = \frac{\partial v_{\pi}(s_0)}{ \partial 
%	\theta}                                                                                                             
%\\
%& = \frac{\partial }{ \partial \theta} \sum_a \pi(s,a)q_\pi 
%(s,a)                                                                                                                  
%\\
%& = \sum_a \frac{\partial \pi(s,a)}{\partial \theta} q_\pi (s,a) + \pi(s,a)  \frac{\partial q_\pi(s,a)}{\partial 
%	\theta}                                                           \\
%& = \sum_a \frac{\partial \pi(s,a)}{\partial \theta} q_\pi (s,a) + \pi(s,a)  \frac{\partial}{\partial \theta} 
%\Big{[}r(s,a) + \sum_{s'} \gamma \mc{P}(s,a,s') v_\pi(s')\Big{]} \\
%& = \sum_a \frac{\partial \pi(s,a)}{\partial \theta} q_\pi (s,a) + \pi(s,a)  \sum_{s'} \gamma \mc{P}(s,a,s') 
%\frac{\partial v_\pi(s')}{\partial \theta}                            \\
%\end{aligned}
%\end{equation}
\begin{equation}
\begin{aligned}
\nabla_\theta J(\theta) &= \nabla_\theta  
v_{\pi}(s_0)                                                                                                         
\\
& = \nabla_\theta \sum_a \pi(s_0,a)q_\pi 
(s_0,a)                                                                                                                 
\\
& = \sum_a \nabla_\theta \pi(s_0,a) q_\pi (s_0,a) + \pi(s_0,a)  \nabla_\theta 
q_\pi(s_0,a)                                                         \\
& = \sum_a \nabla_\theta \pi(s_0,a) q_\pi (s_0,a) + \pi(s_0,a)  \nabla_\theta 
\Big{[}r(s_0,a) + \sum_{s'} \gamma \mc{P}(s_0,a,s') v_\pi(s')\Big{]} \\
& = \sum_a \nabla_\theta \pi(s_0,a) q_\pi (s_0,a) + \pi(s_0,a)  \sum_{s'} \gamma \mc{P}(s_0,a,s') 
\nabla_\theta v_\pi(s')                           \\
\end{aligned}
\end{equation}
We note that we could continue to unroll $\nabla_\theta v_\pi(s')$ on the R.H.S in the same way as 
we have already done. Considering now transitions from starting state $s_0$ to arbitrary state $s$ we therefore find:
\begin{equation}
\begin{aligned}
\nabla_\theta v_{\pi}(s_0) & = \sum_s \sum_{t=0}^{\infty} \gamma^t p(s_t =s| s_0, \pi) 
\sum_a \nabla_\theta \pi(s,a) q_\pi (s,a) \\
\end{aligned}
\end{equation}

where $\sum_{t=0}^{\infty} \gamma^t p(s_t =s| s_0, \pi)$ is the discounted state distribution $\rho^{\pi}(s)$ from 
a fixed starting state $s_0$ (Equation \ref{discounted_state}).
%\tope{it's
%	notationally awkward that $s_0$ then disappears from the succeeding equations. SA: Hmm, I can put it back in if you 
%	like? But its effectively dropped with the $\mb{E}_\pi$ notation anyway right?}
This derivation holds even when there is a distribution over starting states, and gives us the policy gradient theorem:
\begin{equation}
\begin{aligned}
\nabla_\theta J(\theta) = \sum_{s} \rho^{\pi}(s) \sum_a \nabla_{\theta} \pi(s,a) q_{\pi}(s,a)
\end{aligned}
\end{equation}

Using the likelihood ratio trick:

\begin{equation}
\begin{aligned}
\nabla_{\theta} \pi(s,a) & = \pi(s,a) \frac{\nabla_\theta {\pi(s,a)}} {\pi(s,a)} \\
& =\pi(s,a) \nabla_{\theta} \log{\pi(s,a)}
\end{aligned}
\end{equation}

this can be equivalently expressed as:

\begin{equation}
\label{policy_gradient_theorem}
\begin{aligned}
\nabla_\theta J(\theta) & = \sum_{s} \rho^{\pi}(s) \sum_a \pi(s,a) q_{\pi}(s,a) \nabla_{
	\theta} \log{\pi(s,a)}                                                                                \\ &= 
\mb{E}_{\pi} [q_{\pi}(s,a) \nabla_{\theta} \log{\pi(s,a)} ]
\end{aligned}
\end{equation}
%where $q_\pi$ is the true action-value function and:
%
%\begin{equation}
%    \rho^\pi(s) := \int_{\mc{S}} \sum_{t=0}^{\infty}
%    \gamma^{t}p_0(s')p(s_t =s| s', \pi) ds'
%\end{equation}
%is the improper discounted state distribution induced by policy $
%\pi$ starting from an initial state distribution with density $p_0(s')$.

The policy gradient theorem result enables model-free learning as gradients need only be determined for the policy
rather than for properties of the environment. There are a variety of approaches for determining $q_\pi$. If $q_\pi$
is approximated using the sample return (Equation \ref{eqn:return}), this leads to the algorithm known as REINFORCE
\citep{williams1992simple}: 

\begin{equation}
\theta \leftarrow \theta + \alpha  R_t \nabla_\theta \log{\pi (s_t, a_t)}
\end{equation}

As there is no bootstrapping here, this is also known as MC policy
gradient. An alternative approach is to separately approximate $q_\pi$ with a `critic' $q_w$ giving rise to what are 
commonly known as `actor-critic' methods. These introduce two sets of parameter updates; the critic
parameters $w$ are updated to approximate $q_\pi$, and the policy
(actor) parameters $\theta$ are updated according to the policy gradient
as indicated by the critic. The critic itself can be updated according to the TD error. An example of this approach is 
SARSA actor-critic: 

\begin{equation}
\begin{aligned}
w & \leftarrow w + \alpha_1 (r_{t+1}+\gamma q_{w}(s_{t+1},a_{t+1})-q_{w}(s_t,a_t)) \nabla_{w}
q_{w}(s_t,a_t)
\\
\theta & \leftarrow \theta + \alpha_2 q_{w}(s_t,a_t) \nabla_\theta \log{\pi 
	(s_t,a_t)}                           
\end{aligned}
\end{equation}

where different learning rates $\alpha_1$ and $\alpha_2$ may be used for
the actor and the critic.

\subsection{Baselines}
\label{baselines}
Whether we use REINFORCE or an actor-critic based approach to policy gradients, it is possible to reduce the variance 
further by the
introduction of baselines. If this baseline depends only on the state $s$, then we find it introduces no bias:
\begin{equation}
\begin{aligned}
\sum_{s} \rho^{\pi}(s) \sum_a \nabla_{\theta} \pi(s,a)b(s) & = \sum_{s} \rho^{\pi}(s)b(s) \nabla_{\theta} \sum_a \pi(s,
a)                                                                                                                      
\\
& = \sum_{s} \rho^{\pi}(s)b(s) \nabla_{\theta} 1             \\
& = 0
\end{aligned}
\end{equation}

A natural choice for the state-dependent baseline is the value function:
\begin{equation}
\begin{aligned}
\nabla_\theta J(\theta) & = \mb{E}_{\pi} [(q_{\pi}(s,a)-v_\pi(s)) \nabla_{\theta} \log{\pi(s,a)} ] \\ &= \mb{E}_{\pi} 
[A_{\pi}(s,a )
\nabla_{\theta}  \log{\pi(s,a)} ]
\end{aligned}
\end{equation}
where $A_\pi$ is known as the advantage, which may in
some algorithms be approximated directly (rather than approximating both $q_\pi$ and $v_\pi$).

\subsection{Compatible function approximation}
In the general case, our choice to approximate $q_\pi$ with $q_w$ introduces bias such that there are no guarantees of 
convergence to a local optimum. However, in the special case of a compatible function approximator we can introduce no 
bias and take steps in the direction of the true policy gradient. 
This becomes possible when the critic's function approximator reaches a minimum in the mean-squared error:
\begin{equation}
\begin{aligned}
0 & = \mb{E}_{\pi}[\nabla_w (q_{\pi}(s,a)-q_w (s,a))^2]              \\
& =  \mb{E}_{\pi}[(q_{\pi}(s,a)-q_w (s,a))\nabla_w q_w (s,a)]
\end{aligned}
\end{equation}
If we choose $q_w (s,a)$ such that $\nabla_w q_w (s,a) = \nabla_\theta \log{\pi(s,a)}$ we find:

\begin{equation}
\begin{aligned}
\mb{E}_{\pi}[q_{\pi}(s,a)\nabla_\theta \log{\pi(s,a)}] =  \mb{E}_{\pi}[q_w (s,a) \nabla_\theta \log{\pi(s,a)}]
\end{aligned}
\end{equation}
where the L.H.S is equal to the true policy gradient and so our function approximation has introduced no bias. For 
example, if the policy is a Boltzmann policy with a linear combination of features, of the form:

\begin{equation}
\begin{aligned}
\pi(s,a) = \frac{e^{\bm{\theta}^T \bm{\phi}(s,a)}}{\sum_{a'} e^{\bm{\theta}^T \bm{\phi}(s,a')}}
\end{aligned}
\end{equation}

then a compatible value function must be 
linear in the same features as the policy except normalised to zero mean for each state using a subtractive baseline 
\citep{sutton2000policy}. 

\begin{equation}
\begin{aligned}
q_w (s,a) = \bm{w}^T[\bm{\phi}(s,a) - \sum_{a'} \bm{\phi}(s,a') \pi(s,a')]
\end{aligned}
\end{equation}

\subsection{Deterministic policy gradients}

Rather than have a policy specify a probability for certain actions in certain states we can instead have it simply be a function mapping states to actions $\mu_\theta :
\mc{S} \rightarrow \mc{A}$ and, in the case of continuous actions, seek to find the gradient of the objective with respect to the policy parameters. An example of an algorithm which uses such an approach is Deterministic Policy Gradients (DPG) \citep{silver2014deterministic}. The DPG algorithm builds on the deterministic policy gradient theorem:
\begin{equation}
\nabla_\theta J(\theta)= \mb{E}
_{s \sim \rho^\mu}[ \nabla_\theta \mu_{\theta}(s) \nabla_a q_\mu (s,a)|_{a=\mu_{\theta}(s)}].
\end{equation}

where the parameters of the policy are adjusted in an off-policy fashion using an
exploratory behavioural policy (which is a noisy version of the deterministic policy). In practice $q_\mu$ is approximated by the critic $q_w$, which is differentiable in the action and updated using Q-learning:
\begin{equation*}
\begin{aligned}
\delta_t & = r_{t+1} + \gamma q_w (s_{t+1}, \mu_\theta (s_{t+1}))-q_{w} (s_t, a_t) \\ w &\leftarrow w + \alpha_1
\delta_t
\nabla_w q_w (s_t, a_t)                                                         \\
\end{aligned}
\end{equation*} The parameters of the policy are then updated according to:

\begin{equation}
\theta \leftarrow \theta + \alpha_{2} \nabla_\theta \mu_{\theta}(s_t) \nabla_a q_w (s_t,a_t)|_{a=\mu_{
		\theta}(s_t)}
\end{equation}

%  objective $J(\theta)= \mb{E}_{s \sim \rho^
% 	\mu, a \sim \mu_\theta}[r(s,a)]$.

% This discounted state distribution is improper:
%  \begin{equation} %%\rho^\mu(s') := \int_{\mc{S}} \sum_{t=0}^{\infty}\gamma^{t}p_0(s)p(s \rightarrow s', t, \pi) ds
%\rho^\mu(s') := \int_{\mc{S}} \sum_{t=0}^{\infty}\gamma^{t}p_0(s)p(s_t =s'| s, \pi) ds
%\end{equation} %note: I think equation in Silver et al., is slightly wrong with indexing %
%with $p(s \rightarrow s', t, \pi)$ denoting the density at state $s'$ after transitioning for $t$ time steps from 
%state $s$.

\section{Model-based Approaches}

In model-free RL agents learn to take actions directly from experiences, without ever modelling transitions in the environment or reward functions, whereas in model-based RL the agent attempts to learn these. The key benefit is that if the agent can perfectly predict the environment `in its head', then it no longer needs to interact directly with the environment in order to learn an optimal policy. 

% Instead the policy can be learned using dynamic programming or alternatively sampled on simulated experience using model-free RL, which is much more sample efficient. 

% We therefore discuss model learning and sample-based approaches to using models in this section. 

% Whilst so far we have described model-free approaches, an agent can additionally (or alternatively) learn a model and use that to update its value function and learn optimal policies. 

% For some problems, learning
\subsection{Model Learning}
Recall that MDPs are defined in terms of the 5-tuple $\langle \mc{S}$, $\mc{A}$, $\mc{P}$, $r$, $\gamma \rangle$. Although models can be predictions about anything, a natural starting point is to approximate the state transition function $\mc{P}_{\eta} \approx \mc{P}$ and immediate reward function $r_{\eta} \approx r$. We can then use dynamic programming to learn the optimal policy for an approximate MDP $\langle \mc{S}$, $\mc{A}$, $\mc{P}_{\eta}$, $r_{\eta}$, $\gamma \rangle$, the performance of which may be worse than for the true MDP.

Given a fixed set of experiences, a model can be learned using supervised methods. For predicting immediate expected scalar rewards, this is a regression problem whereas for predicting the distribution over next states this a density estimation problem. Given the simplicity of this framing, a range of function approximators may be employed, including neural networks and Gaussian processes.

% However, in many situations an agent will not have access to a model of the environment and even if it did, it may be computationally inefficient to learn by full-width dynamic programming. 

% Whilst model can be used for value function computation, A danger of model learning is introduces another source of approximation error. 

% However some cases underlying model is simple whereas value fucntion is complex. Can learn. Require less interctions with env.

% Whilst interacting with an environment, simply building an accurate model is unlikely to be the sole focus of an agent. Ultimately it wants to optimise its cumulative rewards, and model learning is only a means to an end. In this case, a new exploration-exploitation dilemma occurs for the agent 

% One additional complexity for model learning relates to a similar exploration-exploitation issue found in model-free RL. 

% possibilities are available for model learning, including using neural networks, Gaussian Processes.

%  For stochastic state transitions, models can either output a distribution over states (distribution models) or a probabilistic sample of the next state (sample models). Distribution models can always be used to generate sample models but may be more difficult to learn. 

\subsection{Combining model-free and model-based approaches}

Once a model is learned it can be used for planning. However, in many situations it is computationally infeasible to do the full-width backups of dynamic programming as the state space is too large. Instead, experiences can be sampled from the model and used as data by a model-free algorithm. 

A well known architecture which combines model-based and model-free RL is the Dyna architecture \citep{sutton1991dyna}. Dyna treats samples of simulated and real experience similarly, using both to learn a value function. Simulated experience is generated by the model which is itself learned from real experience. In Dyna, model-free based updates depend on the state the agent is currently in, whereas for the model-based component starting states can be sampled randomly and then rolled forwards using the model to update the value function using e.g. TD learning. 

One potential disadvantage of Dyna is that it does not preferentially treat the state the agent is currently in. In many cases, such as deciding on the next move in chess, it is useful to start all rollouts from the current state (the board position) when choosing the next move. This is known as forward search, where a search tree is built with the current state as the root. Forward based search often uses sample based rollouts rather than full-width ones so as to be computationally tractable and this is known as simulation-based search.

An effective algorithm for simulation-based search is Monte-Carlo Tree search \citep{coulom2007efficient}. It uses the MC return to estimate the action-value function for all nodes in the search tree using the current policy. It then improves the policy, for example by being $\epsilon$-greedy with respect to the new action-value function (or more commonly handling exploration-exploitation using Upper Confidence Trees, see \cite{kocsis2006bandit} for a more detailed discussion). MC Tree Search is equivalent to MC control applied to simulated experience and therefore is guaranteed to converge on the optimal search tree. Instead of using MC control for search it is also possible to use TD-based control, which will increase bias but reduce variance.

Model-based RL is a highly active area of research. Recent advances include MuZero \citep{schrittwieser2020mastering}, which extends model-based predictions to value functions and policies, and Dreamer which plans using latent variable models \citep{hafner2019dream}.

% However, the downside is that the agent may learn an imperfect model of the world which does not match reality. As the agent projects further into the future, errors can grow exponentially making it very challenging to use imperfect models effectively.

% Monte-Carlo Tree Search: Key idea is to preferentially role out simulations from the starting state to extend trajectories that have received high evaluations from earlier simulations. 

\section{Latent variables and partial observability}
\subsection{Latent variable models}
Hidden or `latent' variables
correspond to variables which are not directly observed but nevertheless influence observed variables and thus may be
inferred from observation. In reinforcement learning, it can be beneficial for agents to infer latent variables as
these often provide a simpler and more parsimonious description of the world, enabling better predictions of future
states and thus more effective control.

Latent variable models are common in the field of unsupervised learning. Given data $p(x)$ we may describe a
probability distribution over $x$ according to:

\begin{equation}
p(x; \theta_{x|z}, \theta_z) = \int dz\ p(x | z; \theta_{x|z})p(z; \theta_z)
\end{equation}

where $\theta_{x|z}$ parameterises the conditional distribution $x|z$ and $\theta_z$ parameterises the distribution over
$z$.

Key aims in unsupervised learning include capturing high-dimensional correlations with fewer parameters (as in
probabilistic principal components analysis), generating samples from a data distribution, describing an underlying 
generative process $z$ which describes causes of $x$, and flexibly modelling complex distributions even when the 
underlying components are simple (e.g. belonging to an exponential family).

\subsection{Partially observable Markov decision processes}
\label{background:pomdp}
A partially observable Markov decision process (POMDP)
\citep{kaelbling1998planning} is a generalisation of an MDP in which the
agent cannot directly observe the true state of the system, the dynamics
of which is determined by an MDP. Formally, a POMDP is a 7-tuple
$\langle \mc{S}$, $\mc{A}$, $\mc{P}$, $r$, $\mc{O}$, $\Omega$, $\gamma
\rangle$ where $\mc{S}$ is the set of states, $\mc{A}$ is the set of
actions, $\mc{P} : \mc{ S} \times \mc{A} \times \mc{S} \rightarrow
[0,1]$ is the state transition probability kernel, $r: \mc{S} \times
\mc{A} \times \mc{S} \rightarrow \mb{R}$ is the reward function, $\mc{O}$ is the set
of observations, $\Omega: \mc{S} \times \mc{A} \times \mc{O} \rightarrow
[0,1]$ is the observation probability kernel and $\gamma \in [0,1)$ is
the discount factor. As with MDPs, agents in POMDPs seek to learn a
policy $\pi(s^\alpha_t)$ which maximises some notion of cumulative reward,
commonly $\mb{E}_{ \pi} [\sum_{t=0}^{\infty} \gamma^t r_{t+1}]$. This policy depends on the agent's representation 
of state $s^\alpha_t = f(h_t)$, which is a function of its history. 

One approach to solving POMDPs is by maintaining a belief state over the latent environment state - transitions for which satisfy the Markov property. Maintaining a belief over states only requires knowledge of the previous
belief state, the action taken and the current observation. Beliefs may then be updated according to:

\begin{equation}
b'(s') = \eta \Omega(o'| s', a) \sum_{s \in \mc{S}} \mc{P}(s'|s,a)b(s)
\end{equation}
where $\eta = 1 /
\sum_{s'} \Omega(o'| s', a) \sum_{s \in \mc{S}} \mc{P}(s'|s,a)b(s)$ is a normalising constant.

A Markovian belief state allows a POMDP to be formulated as an MDP where every belief is a state. However, in practice, maintaining belief states in POMDPs will be computationally intractable for any reasonably sized problem. In order to address this, approximate solutions may be used. Alternatively, agents learning using function approximators which condition on the past can construct their own state representations, which may in turn enable relevant aspects of the state to
be approximately Markov.

% Alternatively, single agent RL methods which do
% not explicitly account for partial observability may be applied straightforwardly in the partially-observable setting,
% and have been found to perform well on some problems. In particular, methods which use recurrent neural networks can
% enable agents to construct their own state representations which may in turn enable relevant aspects of the state to
% be approximately Markov.

\section{Deep reinforcement learning}
The policies and value functions used in reinforcement learning can be learned using artificial neural network function 
approximators. When such networks have many layers they are conventionally denoted as `deep', and are typically trained 
on large amounts of data using stochastic gradient descent \citep{lecun2015deep}. The application of deep 
networks in model-free reinforcement learning garnered extensive attention when they were successfully used to learn a 
variety of Atari games from scratch \citep{mnih2013playing}. For the particular problem of learning from pixels a 
convolutional neural network architecture was used  \citep{lecun1998gradient}, which are highly 
effective at extracting useful features from images. They have been extensively used on supervised image classification 
tasks due to their ability to scale to large and complex datasets \citep{lecun2015deep}. 

A deep analysis of deep reinforcement learning (DRL) is beyond the scope of this summary. However we review two key techniques used to overcome the technical challenge of stabilising training.

% The application of deep reinforcement learning (DRL) required the overcoming of major technical challenges. These 
% include the high degree of correlation between states encountered by an RL agent and the non-stationarity of the data 
% distribution as the agent learns. Two approaches to address these issues are `experience replay' and `target networks', 
% which we describe in this section.

% We then explain a DRL algorithm we frequently use in this thesis, known as 
% deep deterministic policy gradients (DDPG), and an approach which can be used to generate differentiable samples from a 
% categorical distribution known as Gumbel Softmax, which we later use with DRL policies. 

\subsection{Experience replay}
As an agent interacts with its environment it receives experiences that
can be used for learning. However, rather than using those experiences immediately, it is possible to store such 
experience in a `replay buffer' and 
sample them at a later point in time for learning. The benefits of such an approach were introduced by  
\cite{mnih2013playing} for their `deep Q-learning' algorithm. At each timestep, this method stores experiences $e_t = 
(s_t, a_t, r_{t+1}, s_{t+1})$ in a replay buffer over many episodes. After sufficient experience has 
been collected, 
Q-learning updates are then applied to randomly sampled experiences from the buffer. This breaks the correlation 
between samples, reducing the variance of updates and the potential to overfit to recent experience. Further improvements to the method can be made by prioritised (as opposed to random) sampling of experiences according to their importance, determined using the temporal-difference error \citep{schaul2015prioritized}.

\subsection{Target networks}
When using temporal difference learning with deep function approximators a common challenge is stability of learning. A 
source of instability arises when the same function approximator is used to evaluate both the value of the current 
state and the value of the target state for the temporal difference update. After such updates, the approximated value 
of both current and target state change (unlike tabular methods), which can lead to a runaway target. To address this, deep RL algorithms often make 
use of a separate target network that remains stable even whilst the standard network is updated. As it is not 
desirable for the target network to diverge too far from the standard network's improved predictions, at fixed 
intervals the parameters of the standard network can be copied to the target network. Alternatively, this transition is 
made more slowly using Polyak averaging:

\begin{equation}
\phi_{targ} \leftarrow \rho \phi_{targ} + (1-\rho) \phi
\end{equation}
where $\phi$ are the parameters of the standard network and $\rho$ is a hyperparameter typically close to 1.

\newpage
\bibliographystyle{plainnat}
\bibliography{thesis_bib}

\end{document}